\documentclass[3p]{elsarticle}
\usepackage{amsopn}
\usepackage{amssymb}
\usepackage{amsmath}

\usepackage{graphicx}
\usepackage{booktabs,pbox}
\usepackage[bookmarks=true%
,bookmarksnumbered=true%
,hypertexnames=false%
,colorlinks=true%
,linkcolor=blue%
,citecolor=blue%
,urlcolor=blue%
]{hyperref}

\usepackage{subfigure} % for subfigures
\usepackage{url,color}
\usepackage{xcolor}
\usepackage[linesnumbered, ruled]{algorithm2e}
\usepackage{algorithmicx}
\usepackage{algpseudocode}% http://ctan.org/pkg/algorithmicx

\newtheorem{definition}{Definition}[section]

\newcommand{\say}[1]{e.g., #1}

\newcommand{\dataset}{{\em Event2012}}

\biboptions{square}
\journal{Journal of Statistics and Management Systems}

\begin{document}
\begin{frontmatter}
\title{Extracting News Events from Microblogs\footnote{This paper was written based on Master's Thesis by \citet{repp:Thesis:2016}.}}

\author{{\O}ystein Repp} 
\author{Heri Ramampiaro}

\address{Dept. of Computer Science\\
Norwegian University of Science and Technology\\
Trondheim, Norway\\
\textbf{Email}: heri@ntnu.no}

% ABSTRACT
\begin{abstract}
Twitter stream has become a large source of information for many people, but the magnitude of tweets and the noisy nature of its content have made harvesting the knowledge from Twitter a challenging task for researchers for a long time. Aiming at overcoming some of the main challenges of extracting the hidden information from tweet streams, this work proposes a new approach for real-time detection of news events from the Twitter stream. We divide our approach into three steps. The first step is to use a neural network or deep learning to detect news-relevant tweets from the stream. The second step is to apply a novel streaming data clustering algorithm to the detected news tweets to form news events. The third and final step is to rank the detected events based on the size of the event clusters and growth speed of the tweet frequencies. We evaluate the proposed system on a large, publicly available corpus of annotated news events from Twitter. As part of the evaluation, we compare our approach with a related state-of-the-art solution. Overall, our experiments and user-based evaluation show that our approach on detecting current (real) news events delivers a state-of-the-art performance.

\end{abstract}
\end{frontmatter}

% SECTIONS THROUGH INPUT FILES
% !TEX root = ./main.tex
\section{Introduction}
\label{sec:introduction}

Due to the proliferation of Smartphones, Social Media, such as Twitter and Instagram, have become one of the most important means of communication. On Twitter alone, there are more than 500 million Twitter status updates (or tweets) every day\footnote{See also \url{http://read.bi/2DdnRFu}.}. Because almost everyone has a smartphone, any person can post messages the moment he/she witnesses or is involved in an event. Still, due to the characteristics of tweets -- including the normal length of a message, its informal and noisy nature, extracting useful peaces of information, including {\em news events}, has not always been easy. Further, people tend to use less distinctive means as humor and sarcasm, abbreviations and emoticons, as well as hashtags which often are concatenations of actual words giving rarely used/non-regular words. For humans, grasping the meaning of tweets is usually straightforward, but automatizing this task has not always been easy. Moreover, the short format of tweets has lead to an enormous Twitter vocabulary, making both sparsity and high dimensionality a big challenge. As a result, traditional text mining techniques normally fail on tweets because they were originally designed for longer texts that are inherently rich in semantics and normally follow strict grammatical rules.    

By overcoming the above challenges, we can harvest all (hidden) information from Twitter streams, thus allowing us to identify and present news events, often reported in real-time by first-hand witnesses. According to \citet{petrovic:CTRN:2013}, even the only 1\% of the public Twitter stream that is made freely available cover around 95\% of all events reported in traditional newswire services. In addition, the Twitter stream contains several other events which are either too small, too local or too domain-specific to be of any interest for any newswire services. Some events may also have value for only a short period of time. As a result, it is crucial to have a system that is able to identify and separate news events from the flood of spam and everyday events reported on Twitter in a timely manner. Journalists, who increasingly adopt social media as a professional tool~\citep{jordan:PMIJ:2013,schifferes:IVNT:2014}, would find such as system very useful; government agents and other organizations, such as  human rights organizations would be interested in early detection of events happening throughout the world~\citep{chen:HRED:2015}; and an average Twitter user would be happy to have a system that enables them to stay up to date on hot news. An extensive study published by the American Press Institute and Twitter shows that nearly 90\% of Twitter users use Twitter for news, and the vast majority of those (74\%) do so on the daily basis~\citep{Rosenstiel2015}. 

The main goal with this work is to develop methods that are able to detect news events from Twitter streams in real time. To achieve this goal we study how to choose features from tweets and how we should represent them when using deep learning to filter news tweets from non-news tweets. Next, we develop a novel clustering method that exploits semantically and temporally similar news-related tweets to form events in real-time. Finally, we investigate an optimal way to rank and present identified news events. 

Again, the aforementioned explosion in the use of social media in the past decade has necessitated developing new methods beyond traditional text analytics methods. Despite the attention to which event detection in Twitter has been subjected, this is still an unsolved task due to the aforementioned characteristics of tweets. Existing approaches have mainly been based on entity, hashtags, and paraphrasing. Our is different from these in that it looks at the entire tweet to achieve good event detection, which means that it does not depend on the existing entity or hashtags in the tweets, thus making the approach more generic and flexible.

The rest of this paper is organized as follows. Section~\ref{sec:preliminaries} describe som preliminaries that our approach builds on. This include establishing the concept of news events. 
Section~\ref{sec:related-work} discusses the related work.  
Section~\ref{sec:approach} explains the method presented in this paper.
Section~\ref{sec:experimental-setup} describes our experimental setup.
Section~\ref{sec:results} presents and discusses the results from our experiments and user evaluation.
Finally, Section~\ref{sec:conclusion} concludes the paper and outline the future work.
%------- 
\section{Preliminaries}
\label{sec:preliminaries}
\subsection{Formal Definition of Event}
\label{sec:preliminaries:definition}
There does not exist any widely agreed definition of "event" as a concept. This makes it difficult to have a common understanding of event detection as a task. This, in turn,  results in comparisons of different detection approaches very hard. It is especially the granularity level of events that differs from one work to another work. Some persons may consider a specific happening a single event (e.g., earth-quake), while others may break the same event into multiple events (e.g, earth quake and tsunami).  This has caused researchers to pose some restrictions and limitations on their systems, making comparison difficult. \citet{McMinn2013} address this problem by creating a large event detection corpus on Twitter data and proposing a more generic definition of events as follows: 

\begin{definition}[Event~\cite{McMinn2013}]
\label{def:event}
\hfill
\begin{enumerate}
	\item An event is a significant thing that happens at some specific time and place.
	\item Something is significant if it may be discussed in the media. For example, you may read a news article or watch a news report about it.
\end{enumerate}
\end{definition}
\noindent

This definition fits well with our objective, and will be used to automatically detect events that can be considered as news events. Note, however, even though a news event must happen at some specific time and place, it does not necessarily mean that tweets referring to an event would mention time and place explicitly. As an example, a tweet might be: "{\ttfamily Police officer shot in bank robbery}". This tweet mentions neither time nor place, but if it refers to an actual episode, that episode is by no doubt a news event. In this paper, the objective is detecting news events in real-time. Therefore, our general assumption is that any mentioned event has news relevance at the time the tweet referring to it is posted. In other words, an event that has already taken place, or that will take place in the future, can still generate a {\em news event} in Twitter today.
Further, we distinguish between \emph{news tweets} and \emph{news events} by the following definition:

\begin{definition}[News tweet and News event] \label{def:newsevent}%\label{def:newstweet}
A {\em news tweet} is a tweet that refers to a specific news event, or in any other way is directly related to such an event as described in Definition~\ref{def:event}.
A {\em news event} is, on the other hand, a group of news tweets that are naturally connected to each other because of temporal and semantic similarities. 
\end{definition}

A more plain explanation of Definition~\ref{def:newsevent} is that it is a group of \emph{news tweets} discussing the same topic at similar times. Hence, any tweets that are not related to news are not considered relevant.
In other words, our focus is on \emph{news events}. Note that whereas news events may refer to any types of news, the event type can unspecified. Further, the time of occurrence for the real-world event that generates an event in Twitter is not important. This is because a sudden increase in the interest it attracts on Twitter indicates that an event my have news value at that moment, and thus constituting a news event. Still, we can assume that Twitter users are more prone to tweets about recent events than old events.

\subsection{Representing Text with Vectors}
\label{sec:preliminaries:representing_text}
An important step before extracting pieces of information, such as events, is a process of transforming unstructured text into a representation that facilitates classification and extraction. In the following, we give an overview of alternative approaches to represent Twitter text with vectors.

\subsubsection{Traditional Language Models}
\label{subsec:tradlm}
A traditional way to represent text data is often as bag-of-words or word-count-based approaches, forming \emph{term vectors}. The vectors consists normally of term weighs computed using their term frequency and inverse document frequency,  i.e., TF-IDF. Unfortunately, due to the aforementioned vocabulary size and word sparseness in social media messages as tweets, this approach is prone to the curse of dimensionality. Some alternative text representations approaches include the named entity vectors~\cite{Kumaran2004} and the mixed vectors, i.e., terms and named entities~\citep{Yang2002}, as well as language models and other probabilistic methods incorporating both content and temporal information~\cite{Ruocco:2012:EAH}. Since, this is beyond the scope of our approach, we refer to the literature (e.g., \cite{Yang2002, Kumaran2004}) for more details.

\subsubsection{Word Embedding Models}
\label{subsec:nlm}
The main problem with word-count-based approaches above is that they ignore the context of the words, e.g., the spatial properties of the words in a sentence or word semantics. During the last few years, researchers have developed a new generation of distributional semantic models, where optimization of the weights in a word vector is performed as a supervised learning task, generally known as word embeddings or context-predicting models. In contrast to traditional methods for construction of context vectors, where the vectors are first created and then re-weighted using various weighting schemes and dimensionality reductions, the context-predicting models aim at optimizing the weights to predict a target word's context. By exploiting the fact that similar words occur in similar contexts, the model learns similar vector representations of similar words~\cite{Baroni2014}.

\paragraph{Word2vec}
One of the most know methods for word embedding is a group of related neural network-based models called Word2vec~\cite{Mikolov2013}.  
The basic idea is of Word2vec is as follows. A Word2vec model uses a large corpus of text as input to produce a vector space, in which each unique word in the corpus are assigned a vector. The model thus positions word vectors in the vector space such that any words sharing common contexts are located close one another in the space.

\citet{Mikolov2013} study two different learning models: \emph{continuous bag-of-words (CBOW)} and \emph{distributed skip-gram}. CBOW model learns to predict a target word in the middle of a symmetric context window, whereas the distributed Skip-gram model learns to predict a target context (i.e. the surrounding words) of a given word. While CBOW is several times faster than Skip-gram and obtains slightly better accuracy for frequent words, Skip-gram works well with small amounts of training data, and represents even rare words or phrases well. %The main reason for this is that with Skip-gram, more training instances can be created from a limited amount of contextual data. With the CBOW method, on the other hand, more data is needed as it depends on surrounding contexts. Both models are so-called "shallow" neural models. They are trained using stochastic gradient descent by back-propagation, with the output word probabilities computed either by \emph{hierarchical softmax} or by \emph{negative sampling}~\cite{Rong2014}.

The main advantage with \emph{Word2vec}, and neural language models in general, is that they are language independent, making them suitable for tweets. 
%
%\todo{[TODO: ADD FIGURES]}

\paragraph{Paragraph Vectors (Paragraph2Vec)}
What can be inferred from the above is that Word2vec is limited to representing only singular words. \citet{Le2014} propose a method called \emph{Paragraph Vector (Paragraph2Vec)}, which adopts a \emph{Word2vec} model to create an unsupervised learning algorithm to provide a fixed-length feature representation from variable-length pieces of text. The main advantage of Paragraph2Vec is that it provides a real-valued dense vector representation of text, and can, for this reason, solve the problem of varying length of text. This, in turn, makes it possible to apply algorithms that require fixed-length features such as SVM or neural network. %Overall, Paragraph2Vec is an interesting method, but more studies are needed to show its viability.

\paragraph{Global Vectors for Word Representation (GloVe)}
%\label{subsec:glove}
\emph{Global Vectors for Word Representation (GloVe)} \citep{glove2014} is another word embedding method. GloVe is a global logbilinear regression model that combines the methods of global matrix factorization and local context window approaches. While Word2vec-based models can be seen as context-predicting models~\cite{Baroni2014}, GloVe is count-based. In general, GloVe has similar performance as Word2vec-based models, though according to \cite{glove2014}, GloVe is 11\% better with respect to accuracy. 
% !TEX root = ./main.tex
\section{Related Work}
\label{sec:related-work}
Detection of events has previously been addressed in the field of information retrieval (IR), but the focus has mainly been on detecting events from conventional media sources, such as RSS feeds from news broadcasters. The Topic Detection and Tracking (TDT) project is possibly the most known example, with which the goal was to produce a system that was able to monitor broadcasted news and produce an alert whenever a new event occurred~\citep{Allan2002}. Most TDT systems have adopted a simple nearest neighbor clustering algorithm, and some reasonably effective systems have been developed. Nevertheless, contents from news broadcasters are different from tweets in that they are normally long, well-written and rich on semantics; while tweets are, as mentioned, very short and often contain large amounts of abbreviated, informal and non-regular words (e.g. hashtags), as well as with grammatical errors and odd sentence structures. Other work has been carried out in the area of detecting events from images, e.g., \cite{ruocco2010event} 

Due to the advance of Internet-based social community, much effort has been put on developing approaches to identify and extract events from different social community resources. Focusing on Twitter, many of the earlier approaches have been concerned with summarizing tweets as part of the event detection~\cite{shao2017efficient,alsaedi2017can}. \citet{Long2011} proposed a language-independent approach for detecting, summarizing and tracking events from tweeter posts.  \citet{ChakrabartiP11} suggested a real time approach to summarize the tweeter posts as events, using a modified variant of the Hidden Markov Model to model the hidden state representation of an event. Other approaches have focused on detecting events in realtime.  In \cite{Watanabe2011}, for instance, the goal was to detect events in realtime from tweet posts by leveraging on their geographical and temporal tags. In \cite{BeckerNG11}, the authors presented a method composed by a clustering step, followed by a classification step to group tweets and separate event clusters from non-event clusters, respectively. Finally, \citet{Sakaki2010} investigated the possibility to detect events such as earthquake using the real time stream of tweet posts as sensors. For this, the authors proposed a specific spatio-temporal model based on Kalman filter to detect such a kind of event.

\citet{Petrovic:2010:SFS} presented a first story detection (FSD) system for Twitter that identified, in a given sequence of tweets, the first tweet to discuss a particular event. More specifically, the authors solved the problem of first story detection by using a variant of the locality sensitive hashing (LSH) technique to determine the novelty of a tweet. The idea is to compare a given tweet with a fixed number of previously encountered  tweets applying a cosinus similarity measure. If a new tweet is considered to represent a new story, it is  assigned to a newly created cluster or a so-called thread. If a tweet, on the other hand, is determined  as  ""not  new", it is assigned to an existing thread containing the nearest  neighbor of that tweet. Event threads are finally ranked based on the combination of the entropy information in a thread and the number of unique user tweets. Further, as mentioned, tweets are inherently noisy, with high lexical variation, making achieving effective FSD performance over social media streams hard. To overcome this problem, \citet{petrovic:2012:UPI} suggested  improving the method presented in \cite{Petrovic:2010:SFS} by using paraphrase expansion, with a paraphrase being a way to express the meaning of a written piece of text with different words. In \cite{petrovic:2012:UPI}'s case, the authors applied the lexical paraphrases extracted from WordNet~\cite{miller:1995:wordnet}, combined with Microsoft Research (MSR) paraphrase tables~\cite{quirk:2004:MMTP} and syntactically constrained paraphrases~\cite{callison:2008:SCPE}. Overall, text expansion using these paraphrases seemed to be effective over newswire, but performance improvements were much smaller when applied over tweets. To further improve the paraphrasing-based FSD approach,  \citet{Moran:2016:EFS} proposed a new technique based on learning word embeddings directly from Twitter data.  More specifically, they suggested to expand tweets with semantically related paraphrases that were identified via automatically mined word embeddings over a background tweet corpus. In this respect, this method is close related to ours in that both apply word embeddings. Note that first story detection is not the same as event detection. They are only related in that event detection must be performed as a part of finding first stories. Moreover, while \cite{Moran:2016:EFS} focused on detecting first story with paraphrasing, our concern is on identifying and classifying news events that are worth presenting to the user.

A work aiming at detecting events in Twitter streams, and thus can be seen as more related to ours, is the work in \cite{McMinn2015}. In their approach, \citet{McMinn2015} used named entities to detect and track real-life events, based on the assumption that named entities -- being  people, places and organizations -- are important building blocks of events.  More specifically, they identified bursty named entities and then used an efficient clustering method to detect and break groups of tweets into individual topics or events. By focusing on tweets containing named entities only, they manage to reduce the amount of data that need to be processed by over 95\%.  \citet{McMinn2015} showed that they were able to detect real life events with high precision while lowering the computational complexity. Nevertheless, since the focus was only entities, this approach seemed to ignore hashtags in tweets. Since hashtag is an integral part of tweets, some important tweets that may constitute events will also be ignored. In our word-embedding approach, we consider every word in tweets to learn the features for event detection in Twitter streams.

Another work worth discussing is the work in \cite{Ritter2012}. In their paper, \citet{Ritter2012} described a system (TwiCal) being an open-domain event-extraction and categorization system for Twitter. The main objective of TwiCal was to extract structured representations of events from Twitter, represented as 4-tuples containing a named entity, an event phrase, a calendar date, and an event type. The focus of TwiCal is on extracting structured representations of events mentioned in tweets. In contrast, our goal is to detect news events, which are of current interest, independent of the time and place of the actual real-life events. \citet{Zhou2014} also focused on extracting structured representations of events. They proposed a Bayesian model, called Latent Event Model (LEM), to extract structured representations of events from social media, represented as named entity, time, location, keywords tuples. This work is interesting because they were able to get good results without using supervised learning, which is beneficial considered the volume of unlabeled data available from the Twitter stream. However, their learning approach does not seem to be done in an online fashion; and to the best of our knowledge, it relies on some pre-configured parameters, including numbers of events. This makes the approach unsuitable for a direct application to the real-time Twitter stream.

\citet{Kunneman2014} proposed an approach, which aims at detecting and separating tweets mentioning significant events from those mentioning everything mundane and insignificant. In particular, their idea is to split each tweet into unigrams and, for each time window, they cluster segments that show a bursty frequency pattern and have similar contents into one cluster. Then, they apply a Hidden Markov Model to uncover the most likely sequence of states for each unigram (bursty/not-bursty). \citet{Agarwal2012} presented a four-step method for extraction of local news events from Twitter, focusing on the specific event types "fire-in-factories" and "labor-strikes". They combine regular expressions based on prior domain knowledge and supervised classification to detect candidate events, which in turn are clustered using LSH.

To summarize, as can be derived from this discussion, the area of event detection from Twitter streams is a highly active research area, and there is a large variation of problems researchers have aimed at solving. Nevertheless, \cite{Petrovic:2010:SFS,Agarwal2012,McMinn2015} seem to be the only works that have focused on new event detection (NED), i.e., the discovery of new events in near real-time. Many existing approaches have focused on retrospective event detection (RED), which typically involves iterative clustering algorithms, or other approaches that are dependent on seeing the full document (tweet) collection, before being able to group and detect events. Only \cite{Petrovic:2010:SFS} explicitly apply their system to the real-time Twitter stream, while other authors experimented with and evaluated their solutions on previously collected tweets. There is also a difference between the approaches regarding whether they address \textit{detection} of events or \textit{extraction} of events. For example, \cite{Ritter2012} and  \cite{Zhou2014} aimed at extraction, which typically includes the extraction of structured representations of events taking place at specific times, as opposed to the other works which simply aim at detecting any mentioned event. 
In conclusion, since our objective is detection of \emph{news events} in the real-time Twitter stream, i.e. unspecified NED, the approach by \citet{McMinn2015} is the work that is the most related to ours.

% !TEX root = ./main.tex
\section{Extraction of News Tweets}
\label{sec:approach}

\begin{figure}[!ht]
\centering
\vspace{3em}
\includegraphics[width=0.9\textwidth]{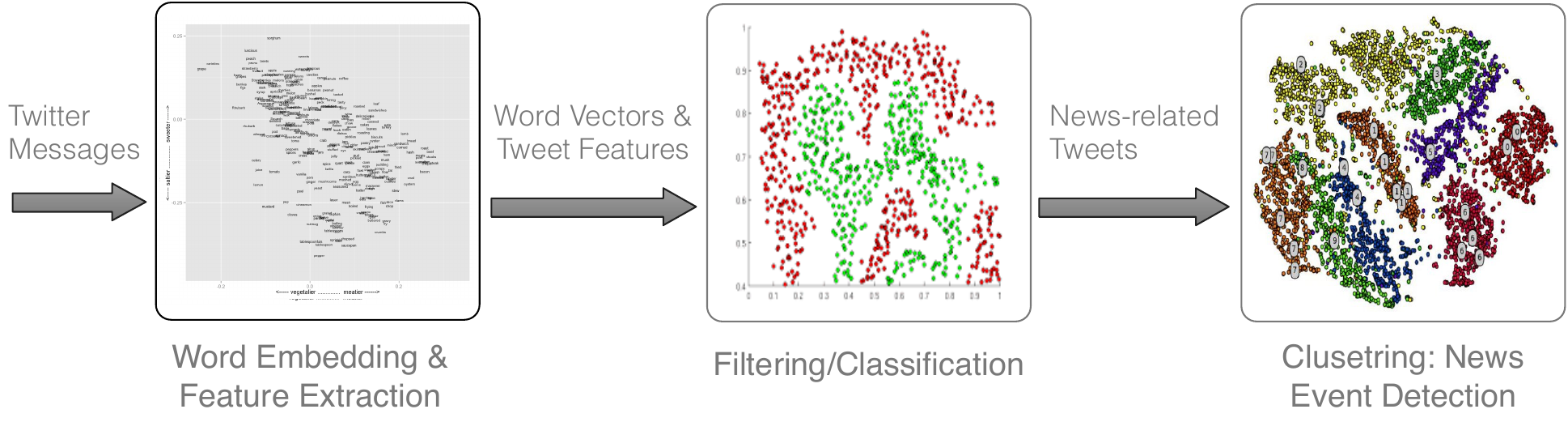}
\caption{Summary of the event detection approach.}
\label{fig:approach-summary}
\end{figure}

Figure~\ref{fig:approach-summary} shows an overview of our approach. As can be seen in this figure we divide our approach into three main steps, elaborated below.

\subsection{Feature Extraction}
\label{sec:approach:feature-extraction}
An important step in the approach is to find an optimal way to represent text. The main goal is to get the best possible for of representation that provide the best classification. For the reason explained in Section~\ref{subsec:tradlm}, traditional methods, such as TF-IDF, are not suitable for our approach due to the afore-mentioned characteristics of tweets. Hence, we focused on applying neural network-based methods. More specifically, we investigated applying (1)~Average Word Vector (AvgW2V), (2)~Paragraph Vectors (D2V), (3)~Average GloVe (AvgGloVe), and (4)~IDF-weighted Word Vector (IDFWV$_{w2v}$) to transform the text from tweets into vectors. 

\subsubsection{Average Word Vector (AvgW2V)}
\label{sec:AvgW2V}
The main idea with Average Word Vector (AvgW2V) is first to generate vectors of words using the neural language model Word2Vec, described in Section~\ref{subsec:nlm}, and then take the average of the vector to represent a text. To do this, we used a model pre-trained by \citet{Godin2015}, consisting of a vocabulary of 3,039,345 words and corresponding word embeddings of 400 dimensions, trained using Google's  Word2Vec toolkit on 400 million pre-processed tweets, collected between 2013-03-01 and 2014-02-28. More specifically, to generate our feature, we first perform pre-processing of the text by filtering out non-English tweets, and replacing user mentions, URLs and numbers with replacement tokens. Next, we generate the word embedding vectors for the words by mapping the words to \citet{Godin2015}'s model. For simplicity,  words that are not present in the model are set as zero-vectors.

\subsubsection{Paragraph Vectors (D2V)}
This method is based on the Paragraph Vector model presented in Section~\ref{subsec:nlm}, using an algorithm based on \emph{doc2vec}. In contrast to AvgW2V the main advantage of this approach is that it is able to capture the ordering of words in the text, and therefore can carry more semantical information. When performing this work, we lacked a pre-trained \emph{doc2vec} model on tweets, so we had to train our own model. More specifically, we performed the training based on the distributed memory algorithm (PV-DM) described in \cite{Le2014}, with 20 full passes over the data (with randomization), and a learning rate starting at 0.025 and decreasing by 0.001 for each epoch. This has resulted in document vectors with dimension 400. The surrounding context for each word is limited to the four closest words.

\subsubsection{Average GloVe (AvgGloVe)}
GloVe is, as mentioned in Section~\ref{subsec:nlm}, very similar to \emph{word2vec}. Similar to word2vec, the authors of GloVe have made available pre-trained word-vectors from 2 billions tweets. The model vocabulary consists of 1.2 million words and the embeddings are available as 25-, 50-, 100-, and 200-dimensional vectors. As with AvgW2V, we applied the average of vectors generated with GloVe to get the vector of each tweets. However, in contrast to the pre-trained model used for \emph{AvgW2V}, with AvgGloVe the words were lowercased at training time and hashtags and user tags were separated from their belonging words. We use the vectors in 200 dimensions, and again words that are not present in the model are set as zero-vectors.

\subsubsection{IDF-Weighted Word Vectors (IDFWV)}
Term frequency--inverse document frequency (TF-IDF) weighting is used in the classic vector space model to incorporate the information value of words in the modeled text. In this method we combine this classical approach with word embeddings. Here, we only used IDF weights, since TF is unlikely to provide any extra value given that tweets are so short (most words occur only once). The IDF weights are calculated from a training set of tweets and used to calculate a weighted mean over the word embeddings for the words in the tweets. Both Word2Vec and GloVe can be used as a basis for the word embeddings.

\subsection{The classification step}
\label{sec:approach:classification}

The main goal with the classification step is to filter out tweets that do not represent news at all. In other words, if a tweet is recognized as an news tweet, the tweet, together with the prediction confidence, is kept as input for the next step, the clustering stage. Otherwise, it is discarded without further action. The classification step comes after the preprocessing  and vector representation step. Taking the vectors as features, we mainly train and use a deep learning model to do the classification task, i.e., to predict whether a tweet is a news tweet or not. In addition to the feature vector generated from textual content of the tweets and the metadata, we investigated using other features as well to optimize the classification step. These include the extracted information, such as number of followers and number of hashtags in the tweet. However, our study showed that these features did not contribute to improve the classification performance. Therefore, we omitted them in the final approach. %These features are summarized in Table~\ref{tab:additional-features}.
%\begin{table}[!ht]
%\centering
%\begin{tabular}{lp{150px}}
%\toprule
%\textbf{Feature name} & \textbf{Description}\\
%\midrule
%\emph{friends-followers ratio} & Number of user's friends divided by number of user's followers.\\ 
%\emph{followers count} & Number of user's followers.\\ 
%\emph{friends count} & Number of user's friends.\\ 
%\emph{statuses count} & Number of user's previous postings.\\
%\emph{hashtags count} & Number of hashtags in tweet.\\
%\emph{mentions count} & Number of user mentions in tweet.\\
%\emph{urls count} & Number of urls in tweet.\\
%\emph{text length} & Number of words in tweet.\\
%\bottomrule
%\end{tabular} 
%\caption[Additional Features]{Additional tweet features} 
%\label{tab:additional-features}
%
%\end{table}

We trained the multilayer neural network classifier in a supervised fashion, labeling input tweets as "{\tt event}" and "{\tt not-event}". In this work, we used Keras\footnote{See \url{http://www.keras.io}.} on top of Theano\footnote{See \url{http://deeplearning.net/software/theano/}.}. A fully connected neural network using {\em dropout}~\cite{Srivastava2014} and {\em Rectified Linear Units} (ReLU) constitutes the architecture of the classifier. Dropout is a regularization technique applied to neural networks to prevent overfitting and to speed up the training by randomly leaving out units from the network throughout the training phase. ReLU, on the other hand, refers to the units the use the computationally cheap Rectified Linear Function as an activation function. Our network consists of four hidden layer containing 400, 400, 200 and 100 neurons, respectively, each of which has a dropout rate of 0.5. Our empirical study has shown that this architecture works well for our purpose, although the exact depth and width of the network did not have any big impact on the performance. %It should also be noted that when using a input layer based on word embeddings, it is basically like adding an extra hidden layer beneath the input layer which then becomes the input of words.

Table~\ref{tab:text_rep_res} shows the impacts of the choice of word vector approaches on the classification accuracy. As we can observe in this table, {\em Average Word Vector (AvgW2V)} is the best method for vector representation. For this reason, we chose this as the main method for generating the feature vectors. 

\begin{table}[!ht]
\small
\centering
\begin{tabular}{lc}
\toprule
\textbf{Method} & \textbf{Mean Accuracy}\\ 

\midrule

$AvgW2V$ & \textbf{0.886} \\
$D2V$ & 0.623 \\
$AvgGloVe$ & 0.684\\
$IDFWV_{w2v}$ & 0.847\\
\bottomrule

\end{tabular} 
\caption{Impacts of the choice of word vector approaches on the classification accuracy.}
\label{tab:text_rep_res}
\end{table}

% --------

\subsection{Event clustering}
\label{sec:approach:event-clustering}
The clustering step is an important step of the detection since it is where we identify to which a news event a news tweet belongs. To do this, we use the same features used in the classification step, but this time we make sure the stopwords and urls are removed, and that the text is normalized by lowercasing all characters. 

Since we are dealing with streaming data, the main requirement is that the clustering is not only effective but also efficient. Using the clustering method proposed \citet{Petrovic2010} as a basis, we propose two alternative extensions of the \citeauthor{Petrovic2010}'s method, one being completely online and the second being a mini-batch variant. We refer to these methods as the {\em online thread-cluster method (oTC)} and the {\em mini-batch thread-cluster method (mbTC)}, respectively. Inspired by \cite{Petrovic2010}, we use the concept "threading" to refer to the fact that detecting news clusters are done by allowing threads of tweets to grow through time, and eventually constituting several {\em news event clusters}.

\subsubsection{Online Thread-Clustering (oTC)}
This method performs online clustering of tweets by assigning the tweet to the same thread as its nearest neighbor if the cosine distance to that neighbor is below a given threshold, $t$. If the distance to its nearest neighbor is above the specified threshold, the tweet is assigned to a new thread. Only the last $w$ tweets are considered when looking up the neighbors. Compared to the approach proposed in \cite{Petrovic2010}, our approach is different in the way we find the nearest neighbors.  While \citet{Petrovic2010} proposed using locality sensitive hashing (LSH) to find nearest neighbors, which is a relaxed nearest neighbor search method, we can afford using true nearest neighbors. This is because the input tweets are already classified as \emph{news tweets}, and thus we do not need to handle the same amount of data. The online thread-clustering method is summarized in Algorithm~\ref{alg:oTC}.

%\IncMargin{1em}
\begin{algorithm}[!ht]
\small
 \SetKwInOut{Input}{input}
 
 \Input{threshold $t$, window size $w$}

 $T\leftarrow \{\}$\;
 
 \While{tweet in stream}{
  $is\_duplicate \leftarrow$ False\;
    
  \eIf{$T$ is empty}{
   $thread\_id(tweet) \leftarrow$ new thread id\;
   }{
   $tweet_{nearest} \leftarrow$ nearest neighbor of $tweet$ in $T$\;
   $d\leftarrow cosine_dist(tweet, tweet_{nearest})$\;
   \uIf{$d \approx 0$}{
    $is\_duplicate \leftarrow$ True\;
    }
    \uElseIf{$d < t$}{
    $thread\_id(tweet) \leftarrow thread\_id(tweet_{nearest})$\;
    }
    \uElse{
    $thread\_id(tweet) \leftarrow$ new thread id\;
    }
  }
  \If{$is\_duplicate$ is False}{
  \lIf{$|T| \geq w$}{remove first tweet from $T$}
  add $tweet$ to $T$\;
  }

 }
 \caption{The online thread clustering algorithm (oTC)}\label{alg:oTC}
\end{algorithm}

\subsubsection{Mini-Batch Thread-Clustering (mbTC)}
The second clustering approach we study is the Mini-Batch Thread-Clustering (mbTC). mbTC is different from oTC in that instead of doing the clustering on the fly, incomming tweets here are handled in batches of a given size, $b$. For every tweet in a new batch, the tweet is assigned the same thread id as its nearest neighbor among the last $w$ previously seen tweets, if the distance between them is below a threshold $t$. Next, another run takes over the tweets in the batch that have \emph{not} already been assigned to a thread. If they have a nearest neighbor within the batch that has already been assigned to a thread and the distance to that neighbor is below $t$, the tweet is assigned to the same thread as that neighbor. Otherwise, we create a new thread to which the tweet is assigned. The details about the mbTC algorithm is shown in Algorithm~\ref{alg:mbTC}.

\begin{algorithm}[!h]
\small
\SetKwInOut{Input}{input}
\Input{threshold $t$, window size $w$, batch size $b$}
$T \leftarrow \{\}$\;
$B \leftarrow \{\}$\;

\While{tweet in stream}{
	add $tweet$ to $B$\;
	\If{$|B| == b$}{
		$duplicates \leftarrow \{\}$\;
		\If{T is not empty}{
			\For{tweet' in B}{
 				$tweet_{nearest}^T \leftarrow$ nearest neighbor of $tweet'$ in $T$\;
 				$d_T\leftarrow dist(tweet', tweet_{nearest}^T)$\;
 				\If{$d_T < t$}{
 					\eIf{$d_T \approx 0$}{
 						add $tweet'$ to $duplicates$\;
 					}{
 						$thread\_id(tweet') \leftarrow thread\_id(tweet_{nearest}^T)$\;							}
 				}
 			}
 		}
 		\For{tweet'' in B that has not been assigned to thread}{
 			$tweet_{nearest}^B \leftarrow$ nearest neighbor of $tweet''$ in $B$\;
			$d_B\leftarrow dist(tweet'', tweet_{nearest}^B)$\;
			\eIf{$d_B < t$ and $tweet_{nearest}^B$ has thread id}{
				\eIf{$d_B \approx 0$}{
					add $tweet''$ to $duplicates$\;		
	 			}{
	 				$thread\_id(tweet'') \leftarrow thread\_id(tweet_{nearest}^B)$\;	
				}
 			}{
 				$thread\_id(tweet'') \leftarrow$ new thread id\;
 			}
 		}
 		\If{$|T| \geq w$}{
 			remove first $|B - duplicates|$ tweets from $T$\;
 		}
 		add all $B - duplicates$ to $T$\;
 		$B \leftarrow \{\}$\;
	}
}
\caption{The mini-batch thread clustering algorithm (mbTC)}\label{alg:mbTC}
\end{algorithm}

%The tweets are then clustered based on the \emph{AvgW2V} vectors in batches of 50 using the \emph{mbTC} algorithm. For every batch, every tweet in it is assigned either to an existing thread (cluster) or to a new thread in which it is the first instance. Every one of these threads correspond to a \emph{news event}. The clustering process is run in parallel with the stream and classifier, and it picks and processes new tweets from the queue of news tweets populated by the classification module. In this way, the clustering process will never affect the system's ability to read the Twitter stream fast enough.

% !TEX root = ./main.tex
\section{Experimental Setup}
\label{sec:experimental-setup}
To evaluate our method we perform both automated evaluation based on existing dataset and a corresponding ground truth. In addition, prove the viability of our approach and the validity of the automated evaluation, we performed user-based evaluation. First, in Section~\ref{sec:dataset}, we describe the dataset we used. Then, in  Section~\ref{sec:evaluation-methodology}, we describe the measures we applied for the evaluation.

\subsection{Dataset}
\label{sec:dataset}
\subsubsection{Dataset with Ground Truth}
\label{sec:dataset:automated}

%\begin{table}[!th]
%\centering
%\begin{tabular}{lrr}
%\toprule
%\textbf{Categories} & \textbf{No. of Events} & \textbf{No. of Tweets}\\ 
%\midrule
%Armed Conflicts \& Attacks & 98 & 8,176 \\ 
%Arts, Culture \& Entertainment & 53 & 6,529\\ 
%Business \& Economy & 23 & 3,448\\ 
%Disasters \& Accidents & 29 & 4,064 \\ 
%Law, Politics \& Scandals  & 140 & 31,261 \\ 
%Miscellaneous & 21 & 3,020 \\ 
%Science \& Technology & 16 & 1,958 \\ 
%Sports & 126 & 24,414\\ 
%\midrule
%Total & 506 & 82,870 \\
%\bottomrule
%\end{tabular} 
%\caption{Event categories and number of events per category, and the number of downloaded labeled tweets in each event category.}
%\label{tab:events-categories}
%\end{table}
Twitter terms of services impose restrictions on re-distribution of tweets, which only allows making available tweet ids and not the tweets themselves. For this reason, obtaining a new up-to-date Twitter dataset is a challenging task, and there is almost no good publicly datasets for event detection tasks.  To evaluate the method proposed in this paper, we extracted a set of labeled and unlabeled tweets from the dataset created by \citet{McMinn2013}, which we from now on refer to as {\em Event2012}. Originally, this collection contained 150K tweet IDs with event labelings. However since the dataset was originally created in 2012, a large portion of the tweets is no longer available, either because the tweets have been deleted already, or the users who originally posted no longer have any valid Twitter accounts. The resulting hydrated dataset contains only  $54 \%$ labeled tweets. In total, our {\em Event2012} ground truth consists of 82,887 labeled tweets covering around 500 different real-life news events. %, for which the distributions are shown in Table~\ref{tab:events-categories}. 

Note that our work is limited to detecting news events without categorizing them into topics. Therefore we merged all the news tweets, except those labeled as {\em Sport}, in {\em Event2012} in to one event class. The reason for not including Sport tweets is because of combination the amount of sport-based tweets and the fact that the sport events  rarely induce breaking news. To summarize, our final ground truth dataset is shown in Table~\ref{tab:tweet_distribution}.

\begin{table}[!htb]

\centering
\begin{tabular}{lc}

\toprule
\textbf{Type} & \textbf{Number of tweets} \\ 
\midrule
Event & 58,456 \\ 
Not-event (incl. sports) & 114,433 \\ 
\midrule
Total & 172,889 \\

\bottomrule

\end{tabular} 
\caption{Distribution of our final dataset after adding irrelevant tweets and merging them with sports tweets.}\label{tab:tweet_distribution}

\end{table}

\subsubsection{Dataset for User-based Evaluation}
\label{sec:dataset:user-based}
Due to issue with the ground truth dataset, we also performed user-based evaluation. For this evaluation, we collected English tweets for a period of 4 days % between 17th of May and 20th of May 
using the public Twitter API. This gave us around 2 mill tweets, which used as an input to our news detection system. We use no threshold on the confidence of tweets detected as news tweets by our classifier, and for the event detection stage (i.e. the clustering stage), only the last 5000 tweets are considered ($w=5000$). The threshold value for assigning a tweet to the same event as its nearest neighbor is set at 0.20. We hypothese that large events (i.e. events with a large number of tweets), as well as rapidly growing events are important news events. We expected our system to return a tremendous amount of news events. This is because all tweets that are classified as \emph{news tweets} (correctly or not) but do not have a nearest neighbor closer than our pre-specified threshold, will be considered new events. To restrict the amount of tweets each person had to evaluate, we picked 100 largest candidate events and 100 events with the fastest increasing positive growth rates in a given time range of four days.
%Hence, to pick candidate news events for our user-based evaluation, we used the 100 largest events and the 100 events with the fastest increasing positive growth rates in a given time range. 
The former lets us pick up large events with a steady growth, whereas the latter ensures that also smaller events can be detected if they are targeted for a sudden rise in interests.

When picking the 100 largest threads we considered only the threads that whose the average classification confidence of all tweets are above 0.85, the average timestamp is within our chosen time window of 24 hours, and the size is larger than 5. To avoid candidate events consisting of tweets produced by Twitter-bots, typically weather reports and traffic updates, we say that at least 85\% of the tweets in an event must be posted by unique users. To reduce the amount of spam and events lacking information (typically "I vote for $<$someone$>$" or "Happy birthday $<$celebrity$>$") we followed the suggestion in \cite{Petrovic2010} and removed events with entropy below a certain threshold. Entropy of an event is computed as
\begin{equation}
Entropy_{event} = -\sum_{i} \dfrac{n_i}{N} \log \dfrac{n_i}{N} ,
\end{equation}
where $n_i$ is the number of times term $i$ appears in the event and $N$ denotes the total number of terms in the event~\cite{Petrovic2010}. We empirically found that setting this threshold at 5 worked well. When picking the final candidate events for the user-based evaluation, we selected 20 events with the highest entropy within our time window.

%\subsection{Evaluation Methodology}
%\label{sec:evaluation-methodology}
%
\subsection{Evaluation Measures}
\label{sec:evaluation-methodology}
Performing a unified evaluation of event detection from microblog streams, such as Twitter, is generally challenging, and finding good measures for quantitative and qualitative performance enabling comparative studies is not a straightforward task. To address the lack of a common evaluation method, \citet{Weiler2015}  argued that a single event detection technique only can be evaluated "against itself", e.g. with respect to different parameter settings. The fact that almost no one publish datasets or source codes, makes also deep comparison with other approaches difficult. Since our focus is on evaluating the ability to detect news events, \emph{recall}, \emph{precision} and {\em f1-score} are suitable evaluation measures. Here, f1-score is given 
\begin{equation}
f1 = \frac{2PR}{(P+R)},
\end{equation}
where $P$ is precision and $R$ denotes recall. For the user-based evaluation, on the other hand, measuring recall was not possible, and therefore we only used precision.

\section{Results}
\label{sec:results}
In this section, we presents the results of our experiments, both from the evaluations with the ground-truth dataset and the user-based evaluation.

\subsection{Ground Truth-based Evaluation}
\label{sec:results:automated}
We aim at performing an evaluation that is reproducible. Since, the \dataset\ dataset is the only publicly available large annotated corpus of news event with Twitter, we used it to train the classifier of our system. To perform the end-to-end evaluation of the system, we split the dataset into distinct and disjoint training and test sets. 

\subsubsection{Comparing the Event Clustering Algorithms}
\label{sec:comparison-clustering}
We ran a series of experiments to decide which of the two algorithms that is best fit for the task of creating news event clusters. To do this, we used the training corpus from \cite{McMinn2013}, but this time we use only the tweets annotated as \emph{news tweets}, assuming this is equivalent to the tweets we get after passing through the classification step. In addition to testing the exact same representation that was found to perform best on the classification task, we try a version using a more aggressive pre-processing step that includes lowercasing the tweets and removing stop-words before the \emph{AvgW2V} vectors are generated. The metrics that we used to assess the cluster quality are \emph{homogeneity}, \emph{completeness}, and \emph{v-measure}~\cite{Rosenberg2007}.

All clusterings in the experiments are done using a window size of 2000, which means that the last 2000 tweets are considered when a new tweet is assigned to a cluster. The test set contains $\sim$54,000 tweets spanning 28 days, meaning that a window size of 2000 corresponds to roughly 25 hours worth of tweets.

Table~\ref{tab:thread_cluster_res} and Table~\ref{tab:cluster_finegrain_res} show the results from comparing the two clustering algorithms. In the first comparison, we are particularly interested in the impacts of applying text preprocessing. As we see in Table~\ref{tab:thread_cluster_res}, both algorithm persistently perform better when the the more aggressive form of pre-processing is used. This result is expected because at this stage, removing stopwords yield higher inter-tweet discrimination degree than keeping the stopwords.

\begin{figure}[h]
\centering
\includegraphics[width=.45\textwidth]{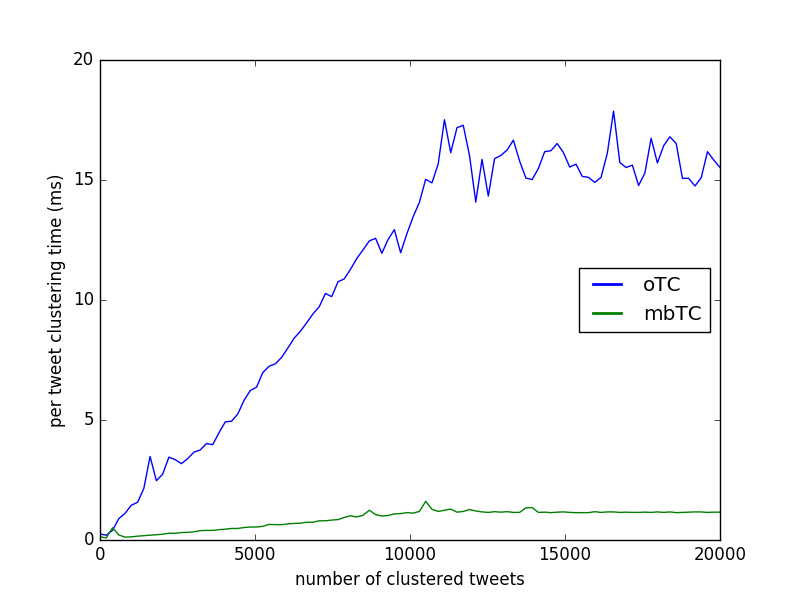}
\caption[Per tweet clustering time for \emph{oTC} and \emph{mbTC}.]{Per tweet clustering time for \emph{oTC} and \emph{mbTC} by number of tweets clustered.}\label{fig:clustertime_graph}
\end{figure}

For \emph{mbTC} there is a trend towards better performance for smaller batch sizes (see Table~\ref{tab:cluster_finegrain_res}). The \emph{mbTC} algorithm using a batch size of 50 does, however, yield better results than the purely online \emph{oTC} algorithm for all threshold values except at $t=0.25$, where \emph{oTC} obtains the best observed score (0.774). We do however see in Table~\ref{tab:cluster_finegrain_res} that we can make the \emph{mbTC} algorithm match the best performance of \emph{oTC} by fine tuning its threshold value. Despite not having a lower complexity than \emph{oTC}, the \emph{mbTC} algorithm is also considerably faster due to less overhead in the implementation (e.g. lists expanded in batches and distances batch-wise calculated using fast matrix operations). The per-tweet clustering time for the two can be seen in Figure~\ref{fig:clustertime_graph}. In this run, the window size was set at 10,000, which explains why the graphs flatten out at 10,000 clustered tweets. As can be observed, \emph{mbTC} is significantly more scalable than  \emph{oTC}, with respect to the number of clustered tweets.
\begin{table}[tbh]
\centering
\begin{tabular}{lcccccccccc}
& \multicolumn{5}{c}{Stopwords included} & \multicolumn{5}{c}{Stopwords excluded}\\
\cmidrule(lr){2-6}\cmidrule(lr){7-11}
\textbf{$t=$} & $.05$ & $.15$ & $.20$ & $.25$ & $.35$ & $.05$ & $.15$ & $.20$ & $.25$ & $.35$\\

\midrule

$oTC$			& .647 & .721 & .681 & .594 & .550 & .646 & .702 & .748 & \textbf{.774} & .665\\
$mbTC_{b=50}$	& .648 & .718 & .589 & .251 & .105 & .648 & .710 & .758 & .763 & .324\\
$mbTC_{b=100}$	& .648 & .709 & .520 & .274 & .136 & .647 & .706 & .749 & .736 & .295\\
$mbTC_{b=200}$	& .647 & .694 & .512 & .301 & .176 & .647 & .699 & .736 & .721 & .305\\
$mbTC_{b=400}$	& .646 & .678 & .497 & .307 & .166 & .646 & .693 & .721 & .698 & .274\\
\bottomrule

\end{tabular} 
\caption{The impacts of preprocessing on the performance in terms of V-Measure for the different clustering methods as function of the threshold value, $t$. The subscript of \emph{mbTC} denotes the batch size that was used. The table includes results from the clustering runs with and without the removal of stopwords.} \label{tab:thread_cluster_res}

\end{table}

\begin{table}[th]
\centering
\begin{tabular}{lcccccccc}
& & \multicolumn{7}{c}{\textbf{Threshold value}}\\
\cmidrule(lr){3-9}
\textbf{Method} & \textbf{Metric} & .22 & .23 & .24 & .25 & .26 & .27 & .28\\

\midrule
				& H	& .958 & .944 & .929 & .910 & .888 & .862 & .833\\
$oTC$			& C & .633 & .648 & .661 & .674 & .687 & .695 & .702\\
				& V & .762 & .768 & .773 & \textbf{.774} & \textbf{.774} & .769 & .762\\
\cmidrule(lr){1-9}
				& H	& .929 & .912 & .884 & .843 & .802 & .762 & .708\\
$mbTC_{b=50}$	& C & .656 & .672 & .686 & .696 & .704 & .717 & .718\\
                & V & .769 & \textbf{.774} & .772 & .763 & .749 & .739 & .713\\
\midrule
\multicolumn{9}{c}{H = homogeneity; C = completeness; V = V-measure}
\end{tabular} 
\caption[]{Performance of \emph{oTC} and \emph{mbTC} by threshold, $t$. The best results are boldfaced. A window size of 2000 was used and stopwords were removed before \emph{AvgW2V} representations were generated.}\label{tab:cluster_finegrain_res}

\end{table}

\subsubsection{Comparison with the State-of-the-art}
\label{sec:comparision}
The approach proposed in \cite{McMinn2015} is, while carrying this work, the only published work that performed an automated evaluation of a similar system with this dataset, we will therefore use this approach as baseline for our evaluation. Note, however, that \citet{McMinn2015} used an unsupervised approach, which let them use the full corpus for testing. For their evaluation, they  included events with at least 75 tweets and they required that at least 5\% or at least 15 tweets in a candidate event must be relevant to a single event from the ground truth for it to be considered detected. In our opinion, however, this threshold is a bit low, because there are many cases in which it would be difficult for a user of the system to identify one specific news event if only a few of the tweets in the candidate event are relevant. In order to facilitate spotting of a news event when presented with the corresponding tweets, it is required that the candidate event contains at least 5 tweets, and that most of the tweets, i.e., at least $80 \%$, in a candidate event must be relevant to the ground truth news event for it to be considered detected. 

Another difference from \cite{McMinn2015}, is that we include all events with 10 tweets or more in the test data and we leave out events larger than 400 tweets. The rationale behind using 10 as a threshold, instead of \cite{McMinn2015}'s 75, is that our test set is already limited as opposed to theirs and we did not want to reduce it further. The reason for cutting off events larger than 400 tweets is that we see that these events in the ground truth dataset typically are news events in a wider definition than ours. As an example, we observed very large events that span several different debates and incidents in the "2012 US Presidential Election". 

For our experiments, we create in total 20 different splits of the ground truth dataset, where the distribution of training--test data is 70-30 (it will vary because we pick events and belonging tweets on random until the test set is at least 30\% of the full corpus). This corresponds to approx. 118K tweets for the training set and 54K tweets for the test set. For each split we train the neural network classifier until convergence using the training set. We then use the corresponding test set to simulate a Twitter stream which delivers event tweets in the order they were posted originally, admixed with non-news tweets. Our system processed this stream until all the tweets in the test set have been classified and possibly assigned to an event. After each run, we analyzed the detected events according to our detection criteria and computed the precision, recall and F1-score for the split. The final result is presented as the mean values of these metrics over all 20 runs.

For the clustering step, we run several iterations to get the optimal value of tweet similarities, i.e., the value for which a tweet would be assigned in a news event cluster. Our empirical study showed that a distance threshold of $0.23$ is the optimal nearest neighbor threshold. The best result from this experiment is presented in Table~\ref{tab:auto_eval_res}.% The window size is set to 2000 tweets.

%\begin{table}[!ht]
%\centering
%\begin{tabular}{lc}
%
%\textbf{Combination of Features} & \textbf{Mean Accuracy}\\ 
%
%\midrule
%
%\emph{Friends-followers ratio + AvgW2V}		&	0.886\\
%\emph{Followers count + AvgW2V}				&	0.886\\
%\emph{Friends count + AvgW2V}				&	0.887\\
%\emph{Statuses count + AvgW2V}				&	0.886\\
%\emph{Hashtags count + AvgW2V}				&	0.888\\
%\emph{Mentions count + AvgW2V}				&	0.886\\
%\emph{URLs count + AvgW2V}					&	0.887\\
%\emph{Text length + AvgW2V}					&	0.888\\
%\emph{All features above + AvgW2V}			&	0.890\\
%\emph{AvgW2V}							&	{\bf 0.889}
%\end{tabular} 
%\caption{The effects of combining other features with \emph{AvgW2V}-vectors.}\label{tab:add_features_res} 
%\end{table}

%-------

\begin{table}[!ht]
\small
\centering
\begin{tabular}{lcc}

~ & \textbf{MMJ\cite{McMinn2015}} & \textbf{Proposed Method} \\ 

\midrule

\textbf{Precision}	& 0.302 (181/586)	& {0.901} (271/300)\\
\textbf{Recall}		& 0.310 (159/506)  	& {0.749} (112/150)\\
\textbf{F1}			& 0.306 			& {0.818} \\

\midrule 
\end{tabular} 
\caption{Results from the work by \citet{McMinn2015} (MMJ\cite{McMinn2015}) and our approach.
The numbers in parentheses are the best runs for MMJ\cite{McMinn2015} based on 75+ tweets, while in ours the numbers are the rounded mean counts over all 20 runs.}
\label{tab:auto_eval_res}

\end{table}

The high F1-score shows that our proposed method has an overall very good ability to detect news events in a stream of tweets. The extremely high average purity as well as the precision score indicate that our $mbTC$ algorithm produces very good events (clusters), although the slightly lower recall rate may indicate that the 0.23 distance threshold for the clustering algorithm is a bit strict or that a window size of 2000 is too low. 

Comparing solutions for event detection in Twitter is, as mentioned already, often problematic due to the lack of datasets and source codes. Nevertheless, our comparison against the approach proposed by \citet{McMinn2015} was not as complete as we intended due to the issues with the availability of Twitter contents. The use of more irrelevant tweets and spam tweets might have caused lower precision than we got, but this would not necessarily have given a more correct picture. For example, actual news events would likely have been detected, but counted as \say{misses} because they were not present in the ground truth. \citeauthor{McMinn2015} accounted for this issue by letting human evaluators judge the relevance of candidate events detected in \dataset\, which obtained a significantly better precision (0.636) than when performing automated evaluation using the original relevance judgments of the corpus (0.302). We also believe that our system would have been able to filter most of any additional spam tweets in the stream. Our news tweet classifier shows good ability to separate irrelevant tweets from news tweets, with a mean precision of 0.907.

On the other hand, our recall value would less likely to degrade even if we increased the amount of spam. The reason for this is that the same events would still have been present in the stream, and our relatively strict threshold for assigning two tweets to the same event would have prevented from including spam tweets. The high average purity of the detected events shows that the proposed method is robust against noises, while still being able to successfully detect them at the 80\% threshold.

\subsection{User-based Evaluation}
\label{sec:results:user-based}
For the reason we described in Section~\ref{sec:dataset:automated}, we have no guarantee about the pureness and completeness of the ground truth data set. Therefor, to make sure that our system over-perform, we decided to evaluate the proposed method with user-based evaluation, allowing users to assess how well the method manages to detect events from current Twitter stream. 

After collecting our dataset as described in Section~\ref{sec:dataset:user-based}, we found 65 candidate events. We presented our users/evaluators with a list of tweets and a corresponding word cloud for each candidate event, and asked them to assess the relevancy of each one. For this, we provided the users with three alternative response options:
\begin{enumerate}
	\item \emph{Yes, the majority of the tweets are relevant to one specific news event}
	\item \emph{No, the majority of the tweets are not relevant to one specific news event, but they are related to a specific news topic}
	\item \emph{No, the majority of the tweets are relevant neither to a specific news event nor a specific news topic}
\end{enumerate}

The reason for the third option, instead of just asking for a yes/no reply to the news event question, is that we wanted to know whether the candidate events not assessed to be news events actually have some news relevance or simply are irrelevant. This would also help us to know the fitness of our chosen granularity level. 

\begin{figure}[!h]
\centering
\vspace{3em}
\includegraphics[width=0.8\textwidth]{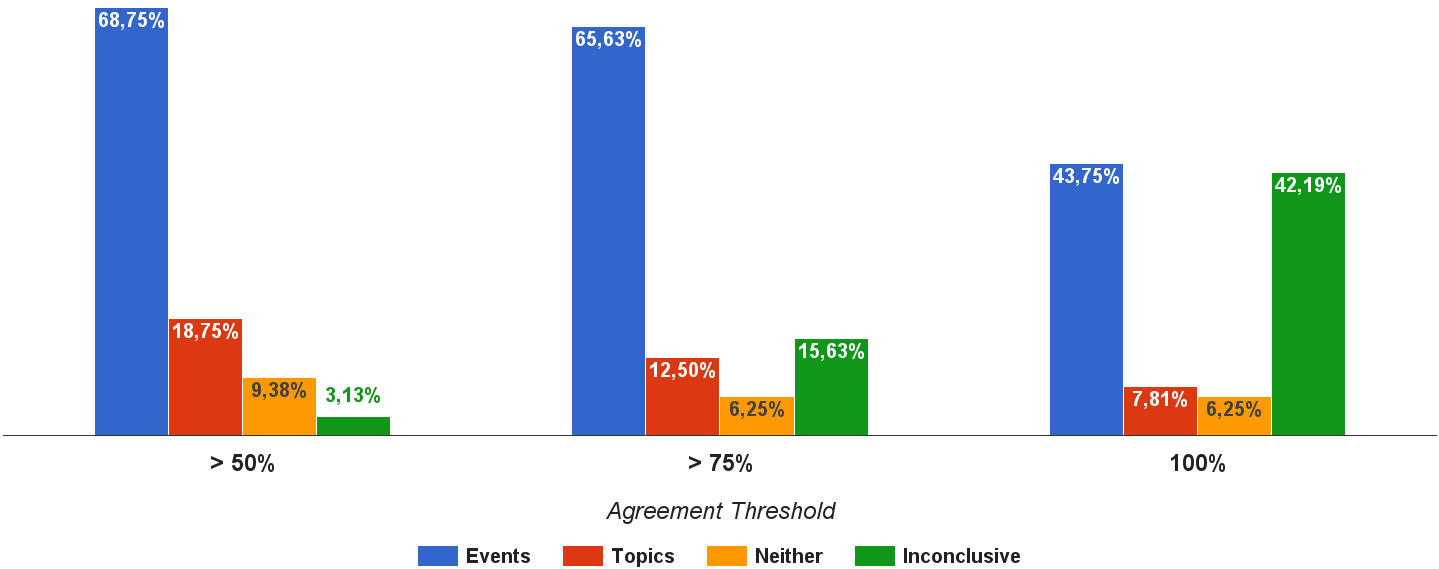}
\caption{The results based on the user evaluation with respect to different annotator agreement levels ($>50\%$ and $>75\%$). The blue bar is for the amount of news events correctly classified by our system according to the user, the red bar is for news topics, and orange bar shows is for the irrelevant tweets. We also include a bar showing the amount of candidate events for which a category none of the user could assess.}
\label{fig:hum_eval}
\end{figure}
Figure~\ref{fig:hum_eval} shows the results from the user evaluation. The results slightly depend on the specified annotator agreement levels. With the majority agreement level, i.e., agreements among more than 50\% of the users, our system manages to correctly identify 68.8\% of the candidate events as being \emph{real news events}. If we increase the agreement level to more than 75\% then we only observe a small change, i.e. 65.6\%. It is worth noting that in the evaluation, we distinguished between "news events" and "news topics", which in some cases could created some confusions for the users. So, by ignoring this difference and looking at all news-related tweets as relevant, the precision value is as high as 90.6\% and 79\%,  respectively, applying the majority and $>75\%$ agreement thresholds. This results are consistent with the our ground truth-based experiments. 

Analyzing the users annotation using kappa statistics, more specifically using Free-marginal multirater kappa~\cite{Randolph2005}, we observed a precision of 71\%, with an inter-annotator agreement 0.76, which indicates a strong agreement and can be regarded as adequate.

This result shows that the approach proposed in this work is capable of detecting real \emph{news events} in the real-time Twitter stream, and that we consider this to be a good result for this specific task. Compared to MMJ\cite{McMinn2015}, this method obtained a precision score of 63.6\% when using crowdsourcing to assess detected events with the \dataset\ dataset. They performed the evaluation on 1210 detected events from the dataset spanning 28 days, which corresponds to around 14 events per day. This is close to our own user-based evaluation where we pick a maximum of 20 candidate events per day. Because of this, we mean that these evaluations are comparable.

%\subsection{Significance of Results}
%The results of the automated evaluation using \dataset\ shows that our system detects news events with a performance that beats an alleged state-of-the-art solution by a great margin. Further does our user-based evaluation show that our system also performs good on today's real-time Twitter stream, and that the precision obtained in the automated evaluation was not greatly overestimated -- especially when considering that the training data was collected in 2012. 

% !TEX root = ./main.tex
\section{Conclusion}
In this paper we proposed a method to extract news events from a stream of tweets. More specifically, we developed a new approach to detect news events from tweets in real-time which consists of three main steps. In the first step, we used a neural network and deep learning-based classification architecture to detect news-relevant tweets from the stream. In the second step we applied a novel streaming data clustering algorithm on the detected news tweets, which formed news events. In the third and final step we proposed method to rank the detected events based on the size of the event clusters and growth speed of the tweet frequencies. We evaluated the proposed system on a large, publicly available corpus of annotated news events from Twitter. As part of the evaluation, we compare our approach with a related state-of-the-art solution. Overall, our experiments and user-based evaluation showed that our approach to detect current (real) news events delivers a state-of-the-art performance.
In our future work, we will further investigate the scalability of our approach, and implement it on in big data platform.
%\todo{Future work: further investigation on scalability, and an end-to-end implementation of the approach using AsterixDB}
\label{sec:conclusion}

% ACKNOWLEDGEMENTS
\section*{Acknowledgement}
This work has been carried out at the Telenor -- NTNU AI-Lab. It is supported by the Dept. of Computer Science, NTNU through the Multi-Source Event Detection (MUSED) project.

\section*{References}
\bibliographystyle{plainnat}
\bibliography{references}  
\end{document}